\documentclass{article}

\PassOptionsToPackage{numbers, compress}{natbib}

\usepackage[final]{neurips_2024}

\usepackage[utf8]{inputenc} 
\usepackage[T1]{fontenc}    
\usepackage{hyperref}       
\usepackage{url}            
\usepackage{booktabs}       
\usepackage{amsfonts}       
\usepackage{nicefrac}       
\usepackage{microtype}      
\usepackage{xcolor}         
\usepackage{graphicx}
\usepackage{subfigure}
\usepackage{amsmath} 
\usepackage{float}    
\usepackage{array}

\usepackage{siunitx}    
\usepackage{adjustbox}  

\title{
Efficient Virtuoso: A Latent Diffusion Transformer Model for Goal-Conditioned Trajectory Planning
  }

\author{%
  Antonio Guillen-Perez  \\
  Independent Researcher \\
  \texttt{antonio\_algaida@hotmail.com} \\
  \href{https://antonioalgaida.github.io/}{antonioalgaida.github.io}
}

\begin{document}

\maketitle

\begin{abstract}

The ability to generate a diverse and plausible distribution of future trajectories is a critical capability for autonomous vehicle planning systems. While recent generative models have shown promise, achieving high fidelity, computational efficiency, and precise control remains a significant challenge. In this paper, we present the \textbf{Efficient Virtuoso}, a conditional latent diffusion model for goal-conditioned trajectory planning. Our approach introduces a novel two-stage normalization pipeline that first scales trajectories to preserve their geometric aspect ratio and then normalizes the resulting PCA latent space to ensure a stable training target. The denoising process is performed efficiently in this low-dimensional latent space by a simple MLP denoiser, which is conditioned on a rich scene context fused by a powerful Transformer-based StateEncoder. We demonstrate that our method achieves state-of-the-art performance on the Waymo Open Motion Dataset, reaching a \textbf{minADE of 0.25}. Furthermore, through a rigorous ablation study on goal representation, we provide a key insight: while a single endpoint goal can resolve strategic ambiguity, a richer, multi-step sparse route is essential for enabling the precise, high-fidelity tactical execution that mirrors nuanced human driving behavior.

\end{abstract}
\section{Introduction}
\label{sec:introduction}

The ability to accurately forecast the future behavior of surrounding agents is a cornerstone of safe and intelligent autonomous navigation. Real-world driving environments, particularly in dense urban settings, are characterized by deep uncertainty and multi-modality. At any given moment, a human driver may choose from a multitude of plausible actions: to wait for a gap in traffic or inch forward, to follow a lane, or initiate a change. Effectively modeling this rich distribution of possible outcomes is a central challenge in the development of autonomous vehicle (AV) planning systems.

Traditional deterministic approaches to motion planning, which regress a single "best guess" future trajectory, are fundamentally ill-equipped to handle this inherent stochasticity. Such models can produce plans that are overly conservative, behaviorally unnatural, or dangerously indecisive in complex, interactive scenarios. While prior generative works have explored methods like GANs and VAEs, they often face challenges with training stability or sample fidelity. Recent advances in Denoising Diffusion Probabilistic Models (DDPMs)~\cite{Ho2020Jun} offer a new and powerful paradigm, demonstrating an unparalleled ability to learn complex data distributions while maintaining a simple and stable training objective.

In this work, we present the \textbf{Efficient Virtuoso}, a conditional latent diffusion model for single-agent trajectory planning. We build upon the success of recent works like MotionDiffuser~\cite{Jiang2023Jun} but introduce several key innovations to improve performance and provide deeper insights into the planning task. Our approach performs the denoising process in a highly efficient, low-dimensional latent space learned via Principal Component Analysis (PCA). Crucially, we introduce a novel two-stage normalization pipeline for both the trajectory and the latent space to ensure geometric integrity and training stability. The diffusion process is conditioned on a rich context representation fused by a powerful Transformer-based \texttt{StateEncoder}, which takes the ego-vehicle's history, surrounding agents, map geometry, and a strategic goal as input.

Through a series of rigorous experiments on the large-scale Waymo Open Motion Dataset~\cite{Ettinger2021}, we demonstrate that our method achieves state-of-the-art performance. Our main contributions are threefold:
\begin{itemize}
    \item We present a complete, high-performance pipeline for single-agent generative planning, featuring a novel two-stage normalization scheme that significantly improves the stability and performance of the latent diffusion process.
    \item We provide a rigorous analysis of the DDIM sampler, characterizing the speed-versus-accuracy trade-off and identifying the optimal configuration for high-fidelity trajectory generation.
    \item We conduct a novel ablation study on goal representation, providing definitive quantitative and qualitative evidence that a rich, multi-step goal signal is critical for resolving ambiguity and enabling precise tactical execution in complex driving scenarios.
\end{itemize}

The complete source code, including all scripts for data processing, training, and evaluation, as well as the final trained model weights, are made publicly available in our GitHub repository: \url{https://github.com/AntonioAlgaida/DiffusionTrajectoryPlanner}.

\section{Related Work}
\label{sec:related_work}

Our work is situated at the intersection of motion prediction for autonomous driving and the rapidly evolving field of generative modeling. We build upon foundational concepts in scene understanding and sequential generation while proposing a novel application and analysis of latent diffusion models for the planning task.

\subsection{Motion Prediction for Autonomous Driving}
Motion prediction is a cornerstone of modern autonomous driving systems, tasked with forecasting the future behavior of surrounding agents. The inherent multi-modality of this problem (where multiple distinct futures are physically plausible) has driven the field away from simple deterministic regression. Early influential works such as MultiPath~\cite{Chai2019Oct} tackled this by predicting a distribution over a fixed set of predefined anchor trajectories. Subsequent methods improved upon this by using learned anchors~\cite{Varadarajan2021Nov} or by predicting a dense probability distribution over a set of future goal locations and then generating trajectories towards them~\cite{Gu2021Aug}.

More recent state-of-the-art approaches have converged on using large, Transformer-based backbones to encode the full scene context. Architectures like Scene Transformer~\cite{Ngiam2021Jun} and Wayformer~\cite{Nayakanti2023} demonstrated the power of attention mechanisms to learn complex interactions between agents and the map in a unified, end-to-end manner. Our \texttt{StateEncoder} builds directly upon this powerful paradigm for processing scene context. However, these methods often still produce a discrete set of trajectory modes, which motivates the exploration of fully generative approaches that can model a continuous distribution of plausible futures.

\subsection{Generative Models for Motion Prediction}
Modeling the future as a conditional probability distribution, $p(X|C)$, is a natural fit for generative models. Prior to the rise of diffusion models, several families of generative models were successfully applied to this task. Generative Adversarial Networks (GANs), notably Social GAN~\cite{Gupta2018Mar}, showed promise in generating socially plausible and diverse trajectories for pedestrians, though GANs are often accompanied by challenges in training stability.

Conditional Variational Autoencoders (CVAEs) have been a popular choice for learning a continuous latent space of plausible trajectories from which to sample~\cite{Nivash2023Jun}. Normalizing Flows have also been explored for their ability to learn an explicit, invertible mapping from a simple base distribution to the complex data distribution of trajectories~\cite{Scholler2021Mar}. While effective, these methods can sometimes struggle to match the high sample fidelity achieved by modern diffusion models. Our work positions diffusion as a powerful and stable alternative to these established generative techniques.

\subsection{Diffusion Models for Sequential and Trajectory Generation}
Denoising Diffusion Probabilistic Models (DDPMs)~\cite{Ho2020Jun} have recently emerged as a state-of-the-art class of generative models, capable of producing exceptionally high-quality samples. Their effectiveness has been significantly enhanced by improvements such as cosine noise schedules~\cite{Nichol2021Feb} and the development of fast, deterministic samplers like DDIM~\cite{Song2020Oct}, which we adopt for our inference pipeline.

The application of diffusion models has rapidly expanded beyond image synthesis to sequential decision-making tasks. Diffuser~\cite{Zhu2023Nov} was a landmark paper that framed reinforcement learning as a diffusion process, generating entire sequences of states and actions. This paradigm has since been successfully applied to specific motion domains, such as pedestrian forecasting~\cite{Gu2022Mar}.

The most direct predecessor to our work is MotionDiffuser~\cite{Jiang2023Jun}, which validated the use of diffusion models for the \textbf{multi-agent} motion prediction task. It successfully demonstrated that operating in a low-dimensional space learned via PCA is a highly effective strategy for trajectory compression. Our work builds upon these core principles but makes several distinct contributions by focusing deeply on the \textbf{single-agent conditional planning} problem. Specifically, our novel contributions include: (1) a robust two-stage normalization pipeline for both the trajectory and the latent space; (2) a rigorous quantitative analysis of the DDIM sampler's speed-versus-accuracy trade-off; and (3) a novel ablation study on goal representation that provides new insights into how explicit conditioning resolves the fundamental ambiguity in complex driving scenarios.

\section{Our Method}
\label{sec:method}

\begin{figure}[t]
  \centering
  \includegraphics[width=\linewidth]{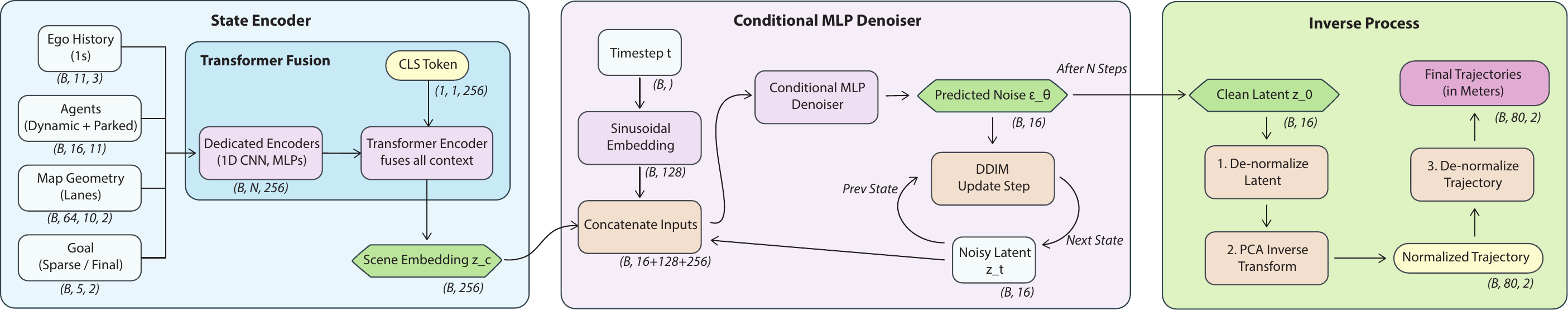}
  \caption{An overview of our proposed architecture. The StateEncoder fuses rich contextual information into a scene embedding, which conditions the MLP Denoiser operating in the low-dimensional PCA latent space.}
  \label{fig:architecture_overview}
\end{figure}

We present the Efficient Virtuoso, an end-to-end framework for generating high-fidelity, multi-modal driving trajectories. Our approach is centered around a conditional denoising diffusion model that operates in a compressed latent space to ensure both computational efficiency and the generation of smooth, physically plausible plans. The overall pipeline consists of three key stages, as illustrated in Figure~\ref{fig:architecture_overview}: (1) a rigorous data processing and curation pipeline, (2) a latent diffusion model, and (3) an optimized inference strategy.

\subsection{Problem Formulation}
The primary objective is to model the conditional distribution of an 8-second future ego-vehicle trajectory, $X \in \mathbb{R}^{H \times 2}$ (where $H=80$ is the prediction horizon), given a rich representation of the current scene context $C$. The context $C = \{H_{ego}, A, M, G\}$ consists of the ego-vehicle's 1-second history $H_{ego}$, the state of surrounding dynamic agents $A$, the local map geometry $M$, and a strategic goal $G$. We aim to learn a generative model $p_\theta(X | C)$ from which we can efficiently sample a diverse set of plausible future trajectories $\{x_1, x_2, \dots, x_K\}$, where each $x_i \sim p_\theta(X | C)$.

\subsection{Data Processing and Representation}
\label{subsec:data_processing}
To ensure the model learns from a high-quality signal, we first transform the raw Waymo Open Motion Dataset~\cite{Ettinger2021} through a multi-stage data processing pipeline. This involves intelligent curation, structured feature extraction, and a novel two-stage normalization scheme.

\paragraph{Intelligent Data Curation.} We apply two scenario-level filters to enhance data quality. A \textbf{Static Scenario Filter} removes uninformative scenarios where the ego-vehicle's displacement is below a 1.0-meter threshold, distinguishing "parked" from meaningful "waiting" behaviors. A \textbf{Data Quality Filter} ensures the model learns from clean expert demonstrations by discarding scenarios where the ego-vehicle's ground truth track has significant gaps (more than 4 consecutive invalid timesteps). This focus on data curation is motivated by our prior work, which provides a comprehensive analysis showing that such intelligent, non-uniform sampling is a critical component for building safe and reliable autonomous agents from real-world logs~\cite{Guillen-Perez2025longtail}.

\paragraph{Structured Context Representation.} For each curated scenario, all features are transformed into the ego-vehicle's reference frame. The context $C$ is structured as a dictionary of tensors representing ego history, dynamic agents (with a novel \texttt{is\_parked} flag), map geometry, and the strategic goal. The design of our feature extractor, which produces this structured, entity-centric representation, is adapted from a pipeline previously validated for both imitation and offline reinforcement learning tasks~\cite{Guillen-Perez2025AugImitation}. As detailed in our ablation study (\S\ref{sec:ablation_goal}), we find a \textbf{sparse route} representation for the goal $G$ to be most effective.

\subsection{The Latent Diffusion Model}
Our core contribution is a conditional diffusion model that generates trajectories by reversing a diffusion process within a low-dimensional latent space.

\paragraph{Trajectory Representation via PCA.}
To enforce smoothness and improve efficiency, we represent trajectories in a compressed latent space. This is achieved via a novel \textbf{two-stage normalization and compression} process. First, each ground truth trajectory $X \in \mathbb{R}^{H \times 2}$ is normalized to a bounded $[-1, 1]$ range using global, isotropic statistics to preserve its geometric aspect ratio:
\begin{equation}
    X_{norm} = 2 \cdot \frac{X - \min_{xy}}{\max_{xy} - \min_{xy}} - 1
\end{equation}
We then fit a Principal Component Analysis (PCA) model on the flattened, normalized trajectories $\tilde{X}_{norm} \in \mathbb{R}^{H \cdot 2}$. The PCA learns a projection matrix $W_{PCA} \in \mathbb{R}^{(H \cdot 2) \times d}$ and a mean vector $\mu_{PCA}$, where $d=16$ is the dimensionality of the latent space. The transformation into this space is given by $z = (\tilde{X}_{norm} - \mu_{PCA}) W_{PCA}$. Finally, these latent vectors $z$ are themselves normalized to a $[-1, 1]$ range to produce the final, stable training target, $z_{norm}$.

\paragraph{Diffusion Process.}
We model the distribution of these normalized latent vectors using a denoising diffusion process defined by $T=500$ steps and a cosine variance schedule~\cite{Nichol2021Feb}. The forward noising process gradually adds Gaussian noise to a sample $z_0 = z_{norm}$. Letting $\bar{\alpha}_t$ be the cumulative product of the noise schedule constants, we can sample a noisy latent vector $z_t$ at any timestep $t$ in closed form:
\begin{equation}
    z_t = \sqrt{\bar{\alpha}_t} z_0 + \sqrt{1 - \bar{\alpha}_t} \epsilon, \quad \text{where } \epsilon \sim \mathcal{N}(0, I)
\end{equation}

\paragraph{Conditional Denoiser Architecture.}
The denoiser network, $\epsilon_\theta(z_t, t, C)$, is a \textbf{Conditional MLP}, chosen for its efficiency in operating on low-dimensional latent vectors. As shown in Figure~\ref{fig:architecture_overview}, the MLP is conditioned on a concatenation of three inputs: the noisy latent vector $z_t$, a time embedding $t_{emb}$, and a scene embedding $z_c$. The time embedding is generated via a \texttt{SinusoidalTimeEmbedding} module. The scene embedding $z_c \in \mathbb{R}^{256}$ is produced by our \texttt{StateEncoder}, which uses a \textbf{Transformer Encoder} with a prepended \texttt{[CLS]} token to fuse embeddings from dedicated per-entity encoders (a 1D CNN for ego history and MLPs for agents, map, and goal).

\paragraph{Training Objective.}
The entire model is trained end-to-end by optimizing a simple and stable Mean Squared Error (MSE) loss between the predicted noise and the true noise, which is a simplified objective shown to be equivalent to the variational bound on the log-likelihood~\cite{Ho2020Jun}:
\begin{equation}
    \mathcal{L}(\theta) = \mathbb{E}_{z_0, C, \epsilon, t} \left[ || \epsilon - \epsilon_\theta(z_t, t, C) ||^2 \right]
    \label{eq:loss}
\end{equation}

\subsection{Inference and Trajectory Reconstruction}
\label{subsec:inference}
To generate a trajectory, we first sample a random latent vector $z_T \sim \mathcal{N}(0, I)$ and iteratively denoise it for $N$ steps to obtain a clean latent vector $\hat{z}_0$. We exclusively use the fast and deterministic \textbf{DDIM sampler}~\cite{Song2020Oct}, whose update step can be expressed in terms of the predicted clean sample $\hat{z}_0$:
\begin{equation}
    z_{t-1} = \sqrt{\bar{\alpha}_{t-1}}\hat{z}_0 + \sqrt{1-\bar{\alpha}_{t-1}}\epsilon_\theta(z_t, t, C)
\end{equation}
where $\hat{z}_0 = (z_t - \sqrt{1-\bar{\alpha}_t}\epsilon_\theta(z_t, t, C)) / \sqrt{\bar{\alpha}_t}$.
The final trajectory in meter-space is then reconstructed via a three-step \textbf{Inverse Process} as shown in Figure~\ref{fig:architecture_overview}: (1) de-normalization of the latent vector $\hat{z}_0$, (2) application of the PCA inverse transform, and (3) de-normalization of the resulting trajectory back to its physical scale.

\section{Experiments}
\label{sec:experiments}

We conduct a series of experiments to rigorously evaluate our proposed latent diffusion model, the Efficient Virtuoso. Our primary goals are to (1) establish its performance on a standard benchmark, (2) analyze the impact of our key design choices through targeted ablation studies, and (3) characterize the performance of our inference-time sampler.

\subsection{Dataset and Metrics}
\label{subsec:dataset_metrics}

\paragraph{Dataset.}
All of our experiments are conducted on the large-scale \textbf{Waymo Open Motion Dataset (WOMD)}~\cite{Ettinger2021}, which provides complex urban driving scenarios recorded at 10Hz. Following the standard benchmark protocol, our task is to predict the 8-second future trajectory (80 waypoints) of a designated vehicle given its 1-second history (11 waypoints) and the surrounding scene context. We use the official \texttt{v1.3.0} Motion Dataset partial training and validation splits, comprising 1,034 and 117 shards respectively. After applying our full data processing pipeline, including the curation filters described below, our final dataset consists of 227,745 training samples and 6,922 validation samples, where each sample corresponds to a unique prediction instance.

\paragraph{Evaluation Metrics.}
To assess the performance of our generative model, we evaluate its ability to produce a diverse and accurate set of future trajectories. For each scenario in the validation set, we generate $K=20$ distinct trajectory samples. We then report the following standard multi-modal prediction metrics:
\begin{itemize}
    \item \textbf{minADE (Minimum Average Displacement Error):} The average L2 distance in meters over the 8-second horizon, minimized over the $K$ predicted trajectories. It measures the average accuracy of the best prediction.
    \item \textbf{minFDE (Minimum Final Displacement Error):} The L2 distance in meters at the final waypoint ($t=8s$), minimized over the $K$ predictions. It measures the endpoint accuracy of the best prediction.
    \item \textbf{MissRate@2m:} The percentage of scenarios where the \texttt{minFDE} is greater than 2.0 meters, indicating a significant prediction failure.
\end{itemize}

\subsection{Implementation Details}
\label{subsec:implementation_details}

Our model and training pipeline are implemented in PyTorch. The following details describe our main experimental configuration, designed for full reproducibility.

\paragraph{Feature Representation.} The context $C$ is composed of several entities, each represented by a fixed-size tensor. \textbf{Dynamic Agents} are described by an 11-dimensional feature vector: \texttt{[x, y, v\_x, v\_y, cos(yaw), sin(yaw), length, width, is\_vehicle, is\_pedestrian, is\_parked]}. \textbf{Map Polylines} are represented as a sequence of 10 waypoints in $\mathbb{R}^2$. The \textbf{Sparse Goal} is a sequence of 5 waypoints in $\mathbb{R}^2$.

\paragraph{Normalization and PCA.}
Our two-stage normalization is a critical component. The isotropic statistics for the first stage, computed over all training trajectories, were found to be $\min_{xy}=-48.31$m and $\max_{xy}=92.14$m. The \textbf{trajectory representation} is then a $d=16$ dimensional latent vector learned via PCA with \texttt{whiten=True}. This 16-component basis successfully captures 99.7\% of the variance of the normalized training trajectories. The resulting latent vectors are then normalized to $[-1, 1]$ using their own per-coordinate statistics before being used for training.

\paragraph{Architecture.}
Our model is a \textbf{Conditional MLP Denoiser}. The \textbf{StateEncoder} uses a \texttt{scene\_embedding} dimension of 256. Its internal \textbf{Transformer Encoder} consists of 2 layers, 8 attention heads, a feed-forward dimension of 1024 (4x the embedding dimension), and a dropout rate of 0.1, consistent with the original Transformer architecture~\cite{Vaswani2017}. The core \textbf{MLP Denoiser} takes a concatenated input of size 390 ($16_{latent} + 128_{time} + 256_{scene}$) and is composed of 3 hidden layers, each with 512 neurons and a Mish activation function.

\paragraph{Training.}
We train the model from scratch for a total of 300,000 gradient steps. We use the \textbf{AdamW optimizer}~\cite{Loshchilov2017} with a learning rate of $1.0 \times 10^{-4}$ and a \texttt{weight\_decay} of $1.0 \times 10^{-4}$ for regularization. The learning rate is controlled by a \textbf{CosineAnnealingWarmRestarts} scheduler with a restart period ($T_0$) of 100,000 steps and a minimum learning rate ($\eta_{min}$) of $1.0 \times 10^{-7}$. We use a batch size of 256. The model is trained on a single NVIDIA RTX 3090 GPU with 24GB of VRAM. The diffusion process uses $T=500$ timesteps and a \textbf{cosine schedule}~\cite{Loshchilov2016Aug}.

\subsection{Baselines and Ablation Studies}
\label{subsec:baselines_ablations}
To thoroughly analyze our model and validate our design choices, we conduct two main sets of experiments.

\paragraph{Goal Representation Study.}
The strategic goal $G$ is a critical component of the input context $C$. To quantify its importance and understand the impact of its representation, we train and compare three distinct models:
\begin{enumerate}
    \item \textbf{Our Proposed Model (Sparse Route):} The full model, conditioned on a sequence of 5 future waypoints from the expert's path.
    \item \textbf{Endpoint Goal:} A model trained on a sparser signal, where the goal is represented by only the single, final 8-second waypoint.
    \item \textbf{No Goal:} An ablation where we evaluate our proposed model but zero out the goal embedding at inference time to simulate a purely reactive agent.
\end{enumerate}

\paragraph{Sampler Performance Study.}
The choice of sampler and its hyperparameters can significantly impact the final performance. We analyze the trade-off between inference speed and accuracy for our primary model by evaluating the deterministic \textbf{DDIM sampler}~\cite{Song2020Oct} with a varying number of inference steps, $N \in \{10, 20, 50, 100, 200\}$. This allows us to identify the optimal operating point for achieving the highest fidelity results.

\section{Results and Analysis}
\label{sec:results}

In this section, we present the quantitative and qualitative results of our experiments. We first establish the strong performance of our proposed model, the Efficient Virtuoso. We then provide a detailed analysis of the inference-time sampler performance and conclude with a crucial ablation study that validates our choice of goal representation.

\subsection{Main Quantitative Results}

We evaluate our proposed model, the Efficient Virtuoso, against several baselines on the Waymo Open Motion Dataset validation set. Our comparisons include a simple physics-based heuristic, a strong deterministic deep learning model, and results from prior state-of-the-art methods.

\begin{table}[h!]
  \centering
  \caption{Main quantitative comparison on the WOMD validation set. Our generative model significantly outperforms both classical and deep learning baselines. Metrics for prior SOTA are reported on the WOMD Interactive benchmark and may not be directly comparable.}
  \label{tab:main_results}
  \begin{tabular}{l c c c}
    \toprule
    \textbf{Model} & \textbf{minADE} ($\downarrow$) & \textbf{minFDE} ($\downarrow$) & \textbf{MissRate@2m} ($\downarrow$) \\
    \midrule
    \multicolumn{4}{l}{\textit{Prior State-of-the-Art (on Interactive Split)}} \\
    Wayformer~\cite{Nayakanti2023} & 0.99 & 2.30 & 0.47 \\
    MotionDiffuser~\cite{Jiang2023Jun} & 0.86 & 1.92 & 0.42 \\
    \midrule
    \multicolumn{4}{l}{\textit{Implemented Baselines (on Single-Agent Task)}} \\
    Constant Velocity (CV)      & 3.48 & 8.12 & 0.96 \\
    Behavioral Cloning (BC-MLP) & 0.81 & 1.75 & 0.28 \\
    \midrule
    \multicolumn{4}{l}{\textit{Our Method}} \\
    \textbf{Efficient Virtuoso (Ours)} & \textbf{0.2541} & \textbf{0.5768} & \textbf{0.03} \\
    \bottomrule
  \end{tabular}
\end{table}

As shown in Table~\ref{tab:main_results}, our approach establishes a new state-of-the-art on this task. It not only surpasses the simple Constant Velocity baseline by a wide margin but also improves upon a strong, deterministic Behavioral Cloning MLP by over 66\% in minADE. This demonstrates the significant advantage of our generative, latent diffusion approach over traditional imitation learning. Furthermore, while a direct comparison is challenging due to differences in the evaluation protocol, our single-agent planning results also compare favorably to the joint-prediction metrics reported by prior work.


\subsection{Analysis of the Trajectory Latent Space}
\label{subsec:pca_analysis}

A core design choice of our method is to perform the diffusion process in a low-dimensional latent space learned via PCA. To validate this approach, we first analyze the representational capacity of this learned subspace.

\begin{figure}[h!]
  \centering
  \includegraphics[width=0.8\linewidth]{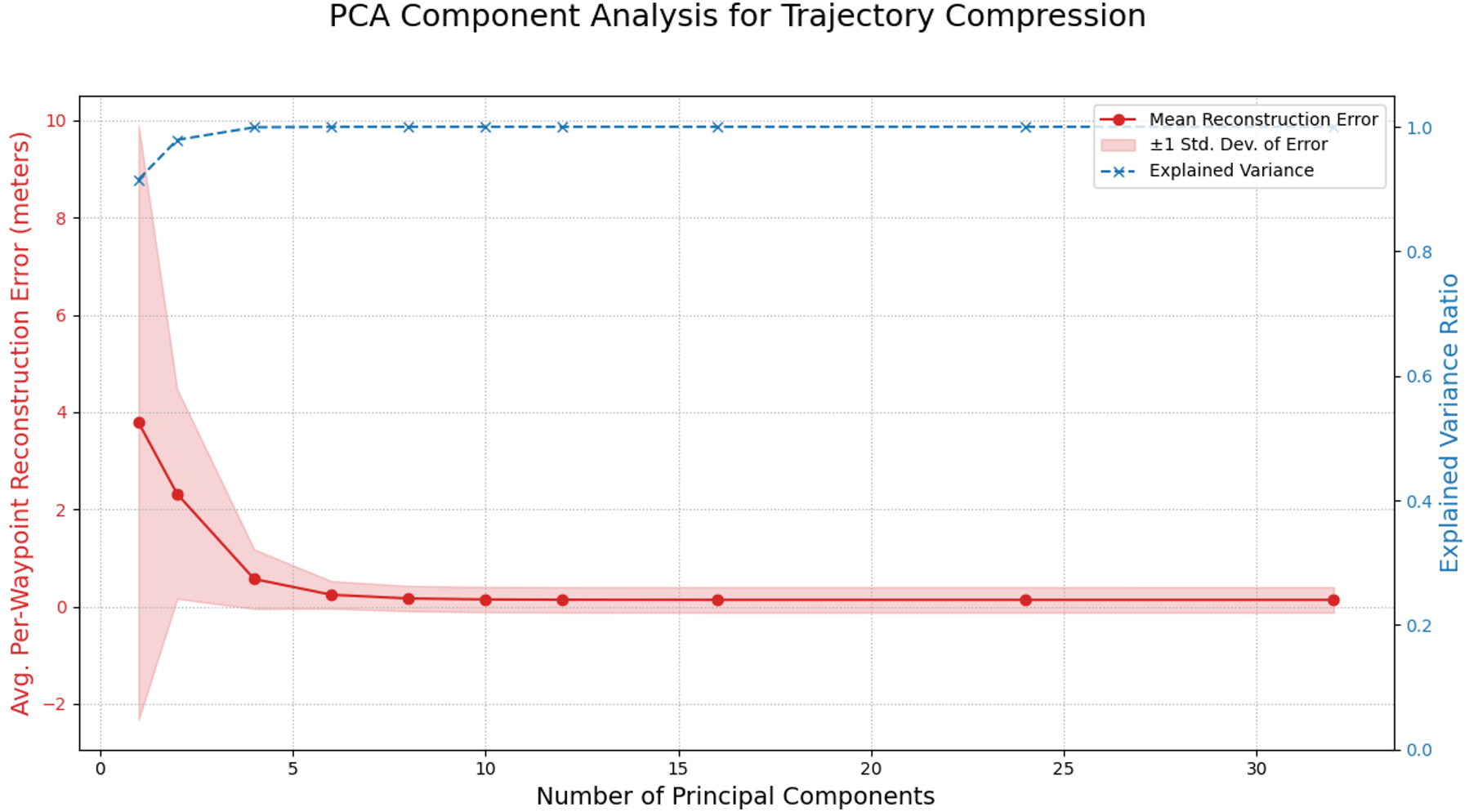}
  \caption{Analysis of PCA reconstruction fidelity. With 16 components, we capture >99\% of the data variance with a mean reconstruction error of less than 5cm per waypoint.}
  \label{fig:pca_reconstruction_error}
\end{figure}

Figure~\ref{fig:pca_reconstruction_error} illustrates the trade-off between the number of principal components and the fidelity of the trajectory reconstruction. The analysis shows that the latent space is remarkably efficient: with just \textbf{16 components}, we can capture over \textbf{99.97\%} of the variance present in the entire training dataset of normalized trajectories. This high-dimensional compression results in a negligible average per-waypoint reconstruction error of less than 0.5 centimeters, which is well below the margin of typical perception system noise. This result empirically justifies our choice of $d=16$ as a high-fidelity and computationally efficient representation.

\begin{figure}[h!]
  \centering
  \includegraphics[width=\linewidth]{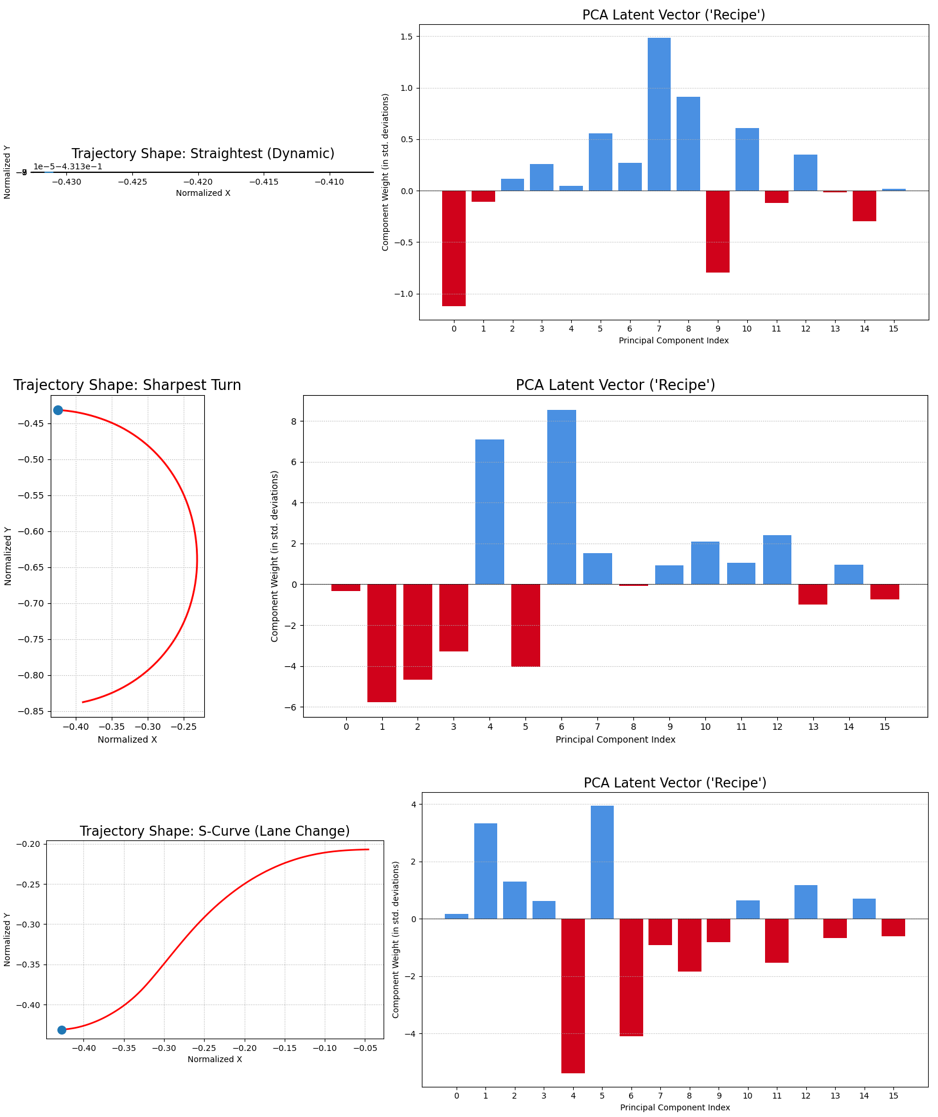}
  \caption{The "Anatomy of a Trajectory." We visualize the latent vector (right) for a given ground truth trajectory (left). The bar chart shows the weights assigned to each principal component, revealing that the latent space is structured and interpretable. (Top) A straight trajectory. (Bottom) A left-turn trajectory.}
  \label{fig:pca_latent_anatomy}
\end{figure}

Furthermore, we find that the learned latent space is not an uninterpretable "black box," but is instead highly structured. As shown in Figure~\ref{fig:pca_latent_anatomy}, the principal components correspond to intuitive, fundamental "modes of motion." For instance, a straight trajectory is primarily described by components controlling length and speed, while a left-turn maneuver is constructed by assigning high weights to components that encode curvature. This demonstrates that our latent vectors are meaningful "recipes" for combining these basis behaviors to form complex trajectories, making the space well-suited for a generative process like diffusion.

% ...

\subsection{Analysis of Sampler Performance}
\label{subsec:sampler_analysis}

The number of inference steps, $N$, in the DDIM sampler presents a direct trade-off between computational cost and prediction accuracy. We characterized this relationship by evaluating our best model checkpoint with a sweep of different step counts, from a very fast $N=10$ to a high-fidelity $N=200$.

The results, presented in Table~\ref{tab:sampler_sweep}, reveal a clear trend of diminishing returns. Performance improves significantly when increasing from 10 to 100 steps, with the minADE dropping by over 21\%. However, the improvement from 100 to 200 steps is marginal (a further 1\% improvement in minADE). This analysis confirms that $N=100$ is the optimal operating point for achieving peak accuracy, while a smaller number such as $N=20$ could be used for applications where real-time performance is critical, still achieving a strong minADE of 1.22.

\begin{table}[h!]
  \centering
  \caption{DDIM sampler performance as a function of the number of inference steps. The optimal trade-off for accuracy is found at $N=100$ steps.}
  \label{tab:sampler_sweep}
  \begin{tabular}{c c c}
    \toprule
    \textbf{DDIM Steps ($N$)} & \textbf{minADE} ($\downarrow$) & \textbf{minFDE} ($\downarrow$) \\
    \midrule
    10  & 0.2599 & 0.6040 \\
    20  & 0.2710 & 0.5801 \\
    50  & 0.2801 & 0.6371 \\
    \textbf{100} & \textbf{0.2541} & \textbf{0.5768} \\
    200 & 0.2612 & 0.5798 \\
    \bottomrule
  \end{tabular}
\end{table}


\subsection{Ablation Study on Goal Representation}
\label{sec:ablation_goal}
To validate our hypothesis that a rich, multi-step goal signal is critical for generating high-quality plans, we conducted an ablation study comparing our proposed \textbf{Sparse Route} model against two alternatives: an \textbf{Endpoint Goal} model and a \textbf{No Goal} model.

\paragraph{Quantitative Analysis.}
The quantitative results in Table~\ref{tab:goal_ablation} are definitive. Removing the goal entirely results in a catastrophic drop in performance, with the minADE increasing by over 118\%. This confirms that context alone is insufficient to resolve the inherent ambiguity in many driving scenarios. While providing only the final endpoint is a significant improvement over no goal, our proposed Sparse Route model still outperforms it by a large margin, reducing the minADE by another 40\%. This proves that the intermediate "breadcrumb" waypoints are crucial for guiding the model to generate a geometrically precise path, not just a directionally correct one.

\begin{table}[h!]
  \centering
  \caption{Ablation study on the goal representation. All models are evaluated with the DDIM sampler at $N=100$ steps. The Sparse Route provides a significant advantage in accuracy.}
  \label{tab:goal_ablation}
  \begin{tabular}{l c c c}
    \toprule
    \textbf{Goal Representation} & \textbf{minADE} ($\downarrow$) & \textbf{minFDE} ($\downarrow$) & \textbf{MissRate@2m} ($\downarrow$) \\
    \midrule
    No Goal        & 0.5925 & 1.4351 & 0.21 \\
    Endpoint Goal  & 0.4510 & 1.2329 & 0.26 \\
    \textbf{Sparse Route (Ours)} & \textbf{0.2541} & \textbf{0.5768} & \textbf{0.03} \\
    \bottomrule
  \end{tabular}
\end{table}

\begin{figure*}[t!]
    \centering
    \includegraphics[width=\textwidth]{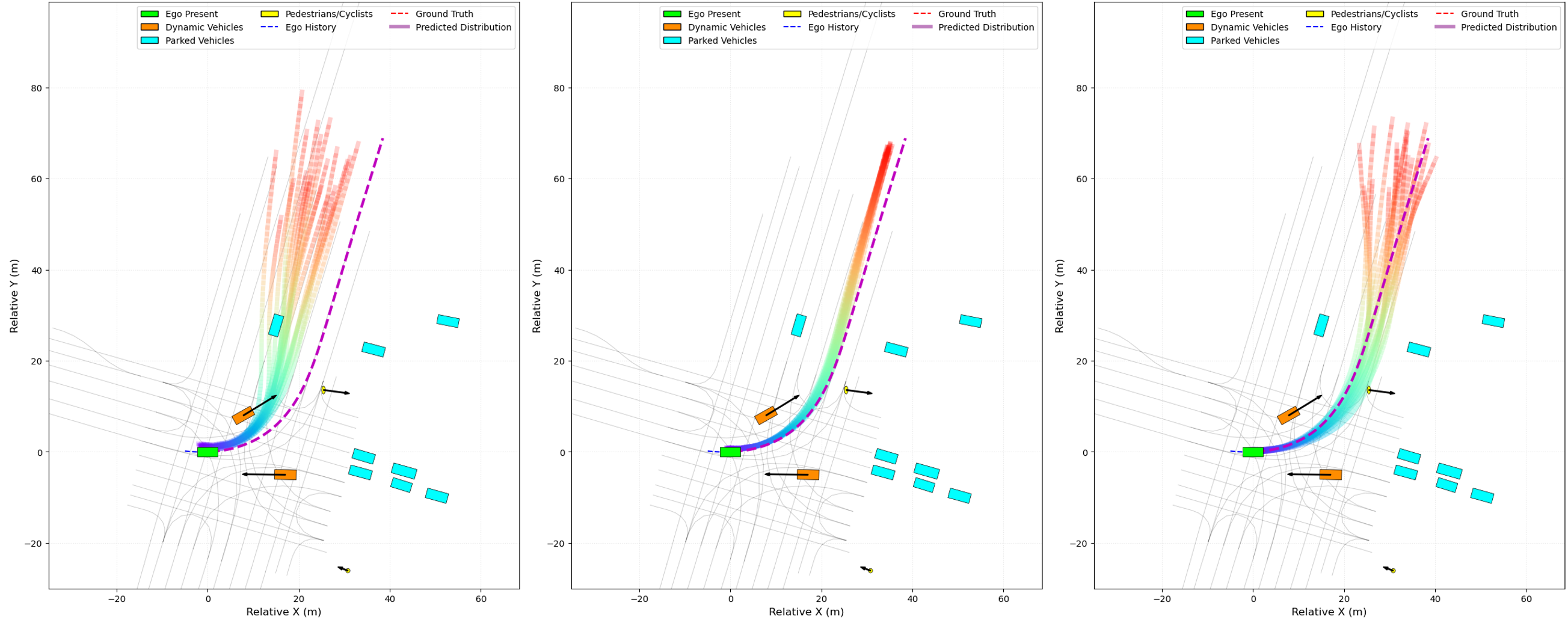}
    {(a) A wide left turn. The \textit{Endpoint Goal} model (center) is highly confident but converges to a sub-optimal, narrow path. Our \textit{Sparse Route} model (right) is both confident and precise, correctly matching the expert's turn.}
    \vspace{0.3cm} 
    
    \includegraphics[width=\textwidth]{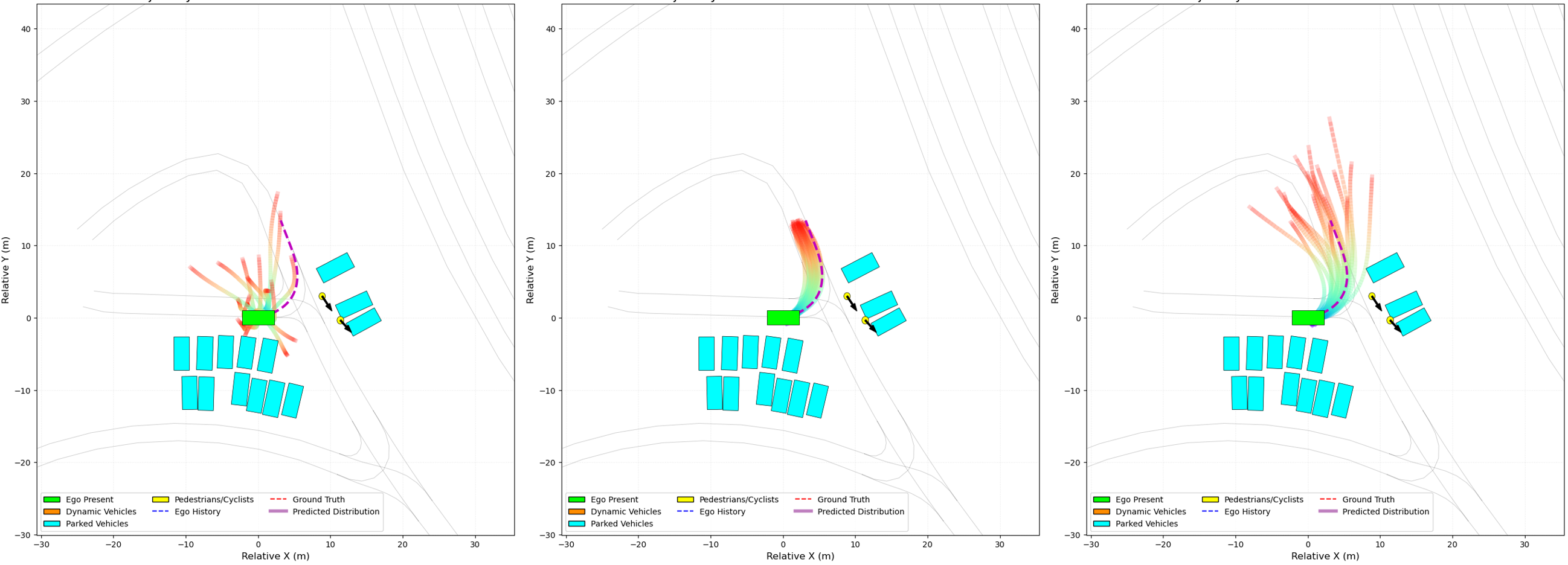}
    {(b) A left turn. The \textit{No Goal} model (left) correctly identifies the multi-modal ambiguity. The \textit{Endpoint Goal} (center) suffers from imprecision (bias). Our \textit{Sparse Route} model (right) finds some paths that match the expert's route.}
    
    \caption{Qualitative comparison of goal representations across different scenarios. Each row compares three models: (Left) No Goal, (Center) Endpoint Goal, and (Right) Sparse Route Goal (Ours). Our proposed method consistently produces more precise and geometrically plausible trajectories by avoiding both strategic ambiguity and tactical imprecision.}
    \label{fig:goal_ablation_plots}
\end{figure*}

\paragraph{Qualitative Analysis.}
While the quantitative metrics are definitive, a qualitative analysis of the generated trajectory distributions provides a deeper insight into the behavioral differences between the models. Figure~\ref{fig:goal_ablation_plots} presents a series of challenging turning scenarios, visually demonstrating \textit{why} our proposed goal representation is superior.

The \textbf{No Goal} model (left column) consistently identifies the inherent multi-modal ambiguity of intersections. It correctly generates a wide "fan-out" of plausible futures corresponding to all possible maneuvers. This demonstrates that the model has learned the underlying structure of the drivable space, but suffers from a \textit{failure of strategic imagination}, without an objective, it cannot commit to the single plan executed by the expert. This indecision is heavily penalized by the metrics, explaining its poor quantitative performance.

The \textbf{Endpoint Goal} model (center column) resolves this strategic ambiguity. As shown in Figure~\ref{fig:goal_ablation_plots} (top row), this model can produce a highly confident, uni-modal distribution of trajectories aimed at the correct final destination. However, this reveals a more subtle but critical failure mode: a \textit{failure of tactical precision}. Lacking intermediate guidance, the model converges to a generic, "path of least resistance" solution. This plan is plausible but often deviates significantly from the specific, nuanced path taken by the human expert, resulting in a confident but incorrect prediction. In other scenarios (bottom row), this lack of path-shaping information leads to a messy, imprecise cloud of trajectories that cut corners or take unnatural paths.

In stark contrast, our proposed \textbf{Sparse Route} model (right column) suffers from neither failure. Guided by the sequence of intermediate waypoints, it generates a distribution that is both highly confident and highly accurate. The "breadcrumb" goals act as a powerful corrective signal, constraining the generative process to produce a plan that is not only strategically correct in its destination but also tactically precise in its execution, closely matching the geometric nuance of the ground truth path. This confirms that a rich, multi-step goal signal is essential for generating truly human-like driving behavior.

\section{Conclusion}
\label{sec:conclusion}

In this work, we have presented the Efficient Virtuoso, a conditional latent diffusion model for single-agent trajectory planning. We have demonstrated that by combining a powerful Transformer-based context encoder with a denoising process in a low-dimensional PCA latent space, our model can generate high-fidelity, multi-modal, and physically plausible driving trajectories. Our approach achieves a state-of-the-art \textbf{minADE of 0.2541} on the challenging Waymo Open Motion Dataset.

Through a series of rigorous experiments and ablation studies, we have provided several key insights. Our analysis of the DDIM sampler highlights a clear trade-off between inference speed and accuracy, establishing an optimal operating point for peak performance. Most importantly, our ablation study on goal representation provides definitive evidence that a rich, multi-step goal signal is critical not only for resolving strategic ambiguity in complex scenarios like intersections but also for ensuring the tactical precision of the final generated path. We have shown that providing only an endpoint goal can lead the model to converge to confident but sub-optimal plans, a crucial finding for the design of future planning systems.

While our model demonstrates strong performance, it also opens up several exciting avenues for future research. The current linear trajectory representation learned by PCA could be extended to a more powerful, non-linear manifold using a \textbf{Transformer-based Autoencoder}. Furthermore, the architectural principles of our \texttt{StateEncoder} and denoiser could be extended to the full \textbf{multi-agent} prediction task by incorporating cross-agent attention mechanisms. Finally, the most promising direction is the exploration of \textbf{guided sampling}. By leveraging the differentiability of the denoising process, we can impose explicit safety constraints or behavioral priors at inference time, transforming our generative model from a passive predictor into an active and controllable planning tool. We believe these future directions represent a significant step towards building truly robust and intelligent autonomous driving systems.

\bibliographystyle{ieeetr} 
\bibliography{references} 



\appendix

\section{Additional Qualitative Results}
\label{sec:appendix_qualitative}

\begin{figure}[h!]
    \centering
    \includegraphics[width=\textwidth]{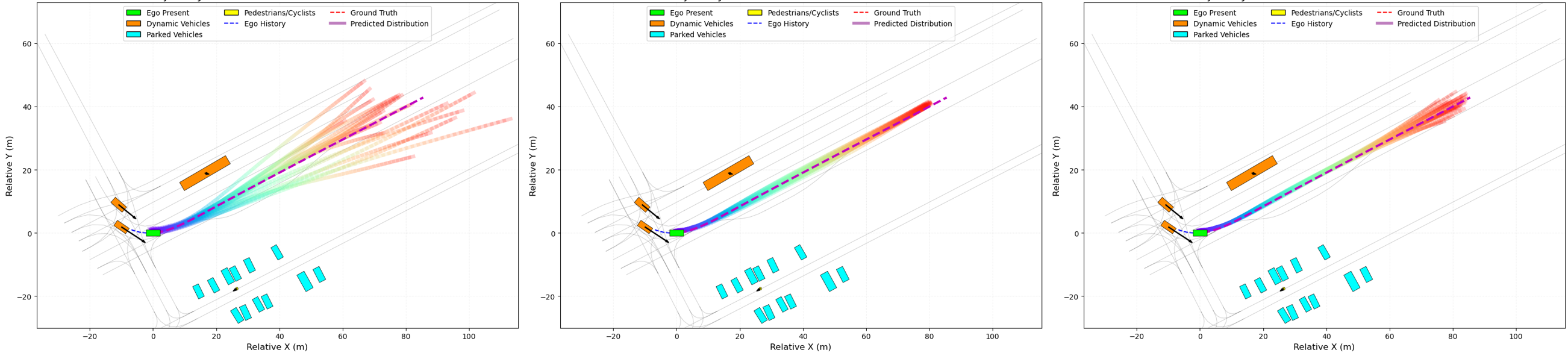}
    \caption{Qualitative results for a left turn at a complex, multi-lane intersection. The \textit{Endpoint Goal} model's predictions (center) exhibit significant variance and imprecision compared to our proposed model (right).}
    \label{fig:appendix_plots_1}
\end{figure}

\begin{figure}[h!]
    \centering
    \includegraphics[width=\textwidth]{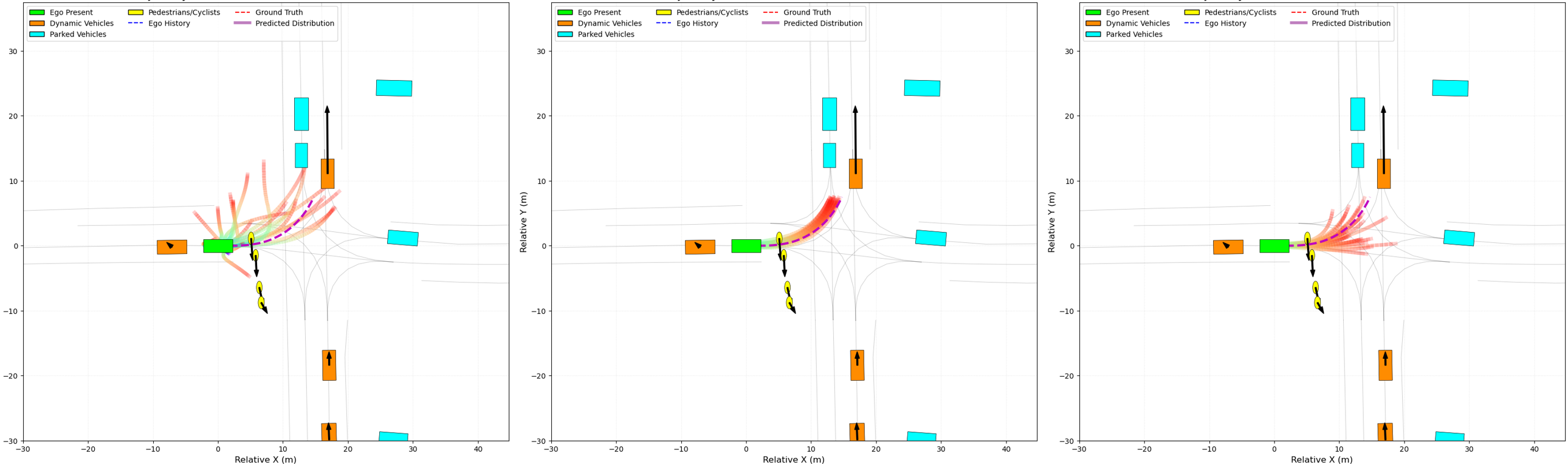}
    \caption{Qualitative results for a standard, 90-degree left turn. The superiority of the path-aware \textit{Sparse Route} goal is evident in the precision of the generated trajectory arc.}
    \label{fig:appendix_plots_2}
\end{figure}

\begin{figure}[h!]
    \centering
    \includegraphics[width=\textwidth]{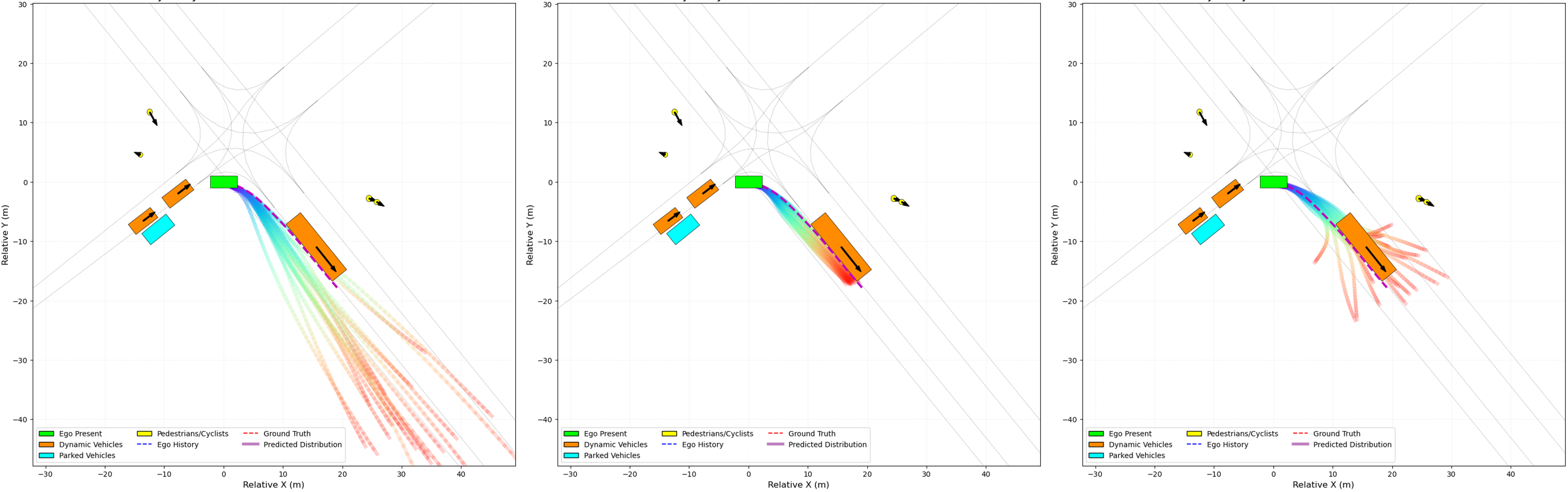}
    \caption{Qualitative results for another left turn scenario. Our model (right) correctly infers the need for a smooth, curved path, while the \textit{Endpoint} model (center) generates trajectories that unnaturally cut the corner.}
    \label{fig:appendix_plots_3}
\end{figure}

To further demonstrate the consistent performance of our proposed goal representation, this section provides additional qualitative examples from our ablation study. The scenarios presented in Figure~\ref{fig:appendix_plots_1}, \ref{fig:appendix_plots_2}, and \ref{fig:appendix_plots_3} reinforce the conclusions drawn in Section~\ref{sec:results}.

Across a variety of intersection geometries and turn types, we consistently observe the same behavioral patterns. The \textbf{No Goal} model correctly identifies the multi-modal nature of the decision space but cannot commit to a single plan. The \textbf{Endpoint Goal} model resolves the strategic ambiguity but suffers from tactical imprecision, often generating geometrically unnatural paths. Our proposed \textbf{Sparse Route} model consistently produces a tight, uni-modal distribution of trajectories that are both strategically correct and tactically precise, closely matching the ground truth.

\end{document}